# Maximum Correntropy Kalman Filter

Badong Chen, *Senior Member, IEEE*, Xi Liu, Haiquan Zhao, José C. Príncipe, *Fellow, IEEE*

*Abstract*—Traditional Kalman filter (KF) is derived under the well-known *minimum mean square error* (MMSE) criterion, which is optimal under Gaussian assumption. However, when the signals are non-Gaussian, especially when the system is disturbed by some heavy-tailed impulsive noises, the performance of KF will deteriorate seriously. To improve the robustness of KF against impulsive noises, we propose in this work a new Kalman filter, called the *maximum correntropy Kalman filter* (MCKF), which adopts the robust *maximum correntropy criterion* (MCC) as the optimality criterion, instead of using the MMSE. Similar to the traditional KF, the state mean and covariance matrix propagation equations are used to give prior estimations of the state and covariance matrix in MCKF. A novel fixed-point algorithm is then used to update the posterior estimations. A sufficient condition that guarantees the convergence of the fixed-point algorithm is given. Illustration examples are presented to demonstrate the effectiveness and robustness of the new algorithm.

*Index Terms*— Kalman Filter, Maximum Correntropy Criterion (MCC), Fixed-Point Algorithm.

## I. INTRODUCTION

Estimation problem has been one of the most important issues from industrial appliances to research areas including signal processing, optimal control, navigation and so on. The actual applications include parameter estimate [27], system identification [28], target tracking [29], simultaneous localization [30] and many others. For linear dynamic systems, the estimation problem is usually solved by Kalman filter (KF), which is, in essence, an adaptive least square error filter that provides an optimal recursive solution [1] [2] [3]. The KF performs very well in Gaussian noises [4]. Nevertheless, its performance is likely to get worse when applied to non-Gaussian situations, especially when the systems are disturbed by impulsive noises. The main reason for this is that KF is based on the well-known *minimum mean square error* (MMSE) criterion, which is sensitive to large outliers and results in deterioration of the robustness of the KF in non-Gaussian noise environments [5].

This work was supported by 973 Program (No. 2015CB351703) and the National Natural Science Foundation of China (No. 61372152).

B. Chen (Corresponding author) and X. Liu are with the School of Electronic and Information Engineering, Xi'an Jiaotong University, Xi'an, China (e-mail: chenbd@mail.xjtu.edu.cn ; lx1102@stu.xjtu.edu.cn).

Haiquan Zhao is with the School of Electrical Engineering, Southwest Jiaotong University, Chengdu, China. (hqzhao@home.swjtu.edu.cn)

J. C. Principe is with the Department of Electrical and Computer Engineering, University of Florida, Gainesville FL 32611 USA, and the School of Electronic and Information Engineering, Xi'an Jiaotong University, Xi'an, China (e-mail: principe@cnel.ufl.edu).

The optimization criteria in *information theoretic learning* (ITL) [6] [7] have gained increasing attention over the past few years, which uses the information theoretic quantities (e.g. entropy) estimated directly from the data instead of the usual second order statistical measures, such as variance and covariance, as the optimization costs. Information theoretic quantities can capture higher-order statistics and offer potentially significant performance improvement in machine learning and signal processing applications. The ITL links information theory, nonparametric estimators, and reproducing kernel Hilbert spaces (RKHS) in a simple and unconventional way. In particular, the *correntropy* as a nonlinear similarity measure in kernel space has its root in Renyi's entropy [8]-[12]. Since correntropy is also a local similarity measure (hence insensitive to outliers), it is naturally a robust cost for machine learning and signal processing [13]-[21]. In supervised learning, such as regression, the problem can be formulated as that of maximizing the correntropy between model output and desired response. This optimization criterion is called in ITL the *maximum correntropy criterion* (MCC) [6] [7]. Recently, the MCC has been successfully used in robust adaptive filtering in impulsive (heavy-tailed) noise environments [6] [9]-[11] [22].

The MCC solution cannot be obtained in closed form even for a simple linear regression problem, so one has to solve it using an iterative update algorithm such as the gradient based methods [9]- [11] [22]. The gradient based methods are simple and widely used. But they depend on a free parameter step-size and usually converge to an optimal solution slowly. The fixed-point iterative algorithm is an alternative efficient way to solve the MCC solution, which involves no step-size and may converge to the solution very fast [6] [24] [25]. A sufficient condition that guarantees the convergence of the fixed-point MCC algorithm was given in [26].

In the present paper, we develop a new Kalman filter, called the *maximum correntropy Kalman filter* (MCKF), based on the MCC and a fixed-point iterative algorithm. Similar to the traditional KF, the MCKF not only retains the state mean propagation process, but also preserves the covariance matrix propagation process. Especially, the new filter has a recursive solution structure and is suitable for online implementation. It is worth noting that in [23], the MCC has been used in hidden state estimation, but it involves no covariance propagation process and is in form not a Kalman filter.

The rest of the paper is organized as follows. In Section II, we briefly introduce the maximum correntropy criterion and Kalman filter. In Section III, we derive the MCKF algorithm and give the computational complexity and convergence analysis. Simulation results are then provided in Section IV to show the excellent performance of the MCKF. Finally, conclusion is given in Section V.

## II. Preliminaries

### A. Maximum Correntropy Criterion

Correntropy is a generalized similarity measure between two random variables. Given two random variables $X, Y \in \mathbb{R}$ with joint distribution function $\mathbf{F}_{XY}(x, y)$, correntropy is defined by

$$V(X,Y) = \mathrm{E}[\kappa(X,Y)] = \int \kappa(x,y) d\mathbf{F}_{XY}(x,y) \qquad (1)$$

where E denotes the expectation operator, and $\kappa(\cdot, \cdot)$ is a shift-invariant Mercer Kernel. In this paper, without mentioned otherwise the kernel function is the Gaussian Kernel given by

$$\kappa(x, y) = G_\sigma(e) = \exp\left(-\frac{e^2}{2\sigma^2}\right) \qquad (2)$$

where $e = x - y$, and $\sigma > 0$ stands for the kernel bandwidth.

In most practical situations, however, only limited number of data are available and the joint distribution $\mathbf{F}_{XY}$ is usually unknown. In these cases, one can estimate the correntropy using a sample mean estimator:

$$\hat{V}(X,Y) = \frac{1}{N}\sum_{i=1}^{N} G_\sigma(e(i)) \qquad (3)$$

where $e(i) = x(i) - y(i)$, $\{x(i), y(i)\}_{i=1}^{N}$ are $N$ samples drawn from $\mathbf{F}_{XY}$.

Taking Taylor series expansion of the Gaussian kernel, we have

$$V(X,Y) = \sum_{n=0}^{\infty} \frac{(-1)^n}{2^n \sigma^{2n} n!} \mathrm{E}\left[(X-Y)^{2n}\right] \qquad (4)$$

As one can see, the correntropy is a weighted sum of all even order moments of the random variable $X - Y$. The kernel bandwidth appears as a parameter weighting the second order and higher order moments. With a very large $\sigma$ (compared to the dynamic range of the data), the correntropy will be dominated by the second order moment.

Given a sequence of error data $\{e(i)\}_{i=1}^{N}$, the cost function of MCC is given by

$$J_{MCC} = \frac{1}{N}\sum_{i=1}^{N} G_\sigma(e(i)) \qquad (5)$$

Suppose the goal is to learn a parameter vector $W$ of an adaptive model, and let $x(i)$ and $y(i)$ denote, respectively, the model output and the desired response. The MCC based learning can be formulated as the following optimization problem:

$$\hat{W} = \arg\max_{W \in \Omega} \frac{1}{N}\sum_{i=1}^{N} G_\sigma(e(i)) \qquad (6)$$

where $\hat{W}$ denotes the optimal solution, and $\Omega$ denotes a feasible set of parameter.

### B. Kalman Filter

Kalman filter provides a powerful tool to deal with state estimation of linear systems, which is an optimal estimator under linear and Gaussian assumptions.

Consider a linear system described by the following state and measurement equations:

$$\mathbf{x}(k) = \mathbf{F}(k-1)\mathbf{x}(k-1) + \mathbf{q}(k-1), \qquad (7)$$
$$\mathbf{y}(k) = \mathbf{H}(k)\mathbf{x}(k) + \mathbf{r}(k). \qquad (8)$$

where $\mathbf{x}(k) \in \mathbb{R}^n$ denotes the $n$-dimensional state vector, $\mathbf{y}(k) \in \mathbb{R}^m$ represents the $m$-dimensional measurement vector at instant $k$. $\mathbf{F}$ and $\mathbf{H}$ stand for, respectively, the system matrix (or state transition matrix) and observation matrix. $\mathbf{q}(k-1)$ and $\mathbf{r}(k)$ are mutually uncorrelated process noise and measurement noise, respectively, with zero mean and covariance matrices

$$\mathrm{E}\left[\mathbf{q}(k-1)\mathbf{q}^T(k-1)\right] = \mathbf{Q}(k-1),\ \mathrm{E}\left[\mathbf{r}(k)\mathbf{r}^T(k)\right] = \mathbf{R}(k) \qquad (9)$$

In general, Kalman filter includes the following two steps:

*1) Predict*

The prior mean and covariance matrix are given by

$$\hat{\mathbf{x}}(k|k-1) = \mathbf{F}(k-1)\hat{\mathbf{x}}(k-1|k-1), \qquad (10)$$
$$\mathbf{P}(k|k-1) = \mathbf{F}(k-1)\mathbf{P}(k-1|k-1)\mathbf{F}^T(k-1) + \mathbf{Q}(k-1). \qquad (11)$$

*2) Update*

The Kalman filter gain is computed as

$$\mathbf{K}(k) = \mathbf{P}(k|k-1)\mathbf{H}^T(k)\left(\mathbf{H}(k)\mathbf{P}(k|k-1)\mathbf{H}^T(k) + \mathbf{R}(k)\right)^{-1} \qquad (12)$$

The posterior state is equal to the prior state plus the innovation weighted by the KF gain,

$$\hat{\mathbf{x}}(k|k) = \hat{\mathbf{x}}(k|k-1) + \mathbf{K}(k)\left(\mathbf{y}(k) - \mathbf{H}(k)\hat{\mathbf{x}}(k|k-1)\right) \qquad (13)$$

Additionally, the posterior covariance is recursively updated as follows:

$$\mathbf{P}(k|k) = (\mathbf{I} - \mathbf{K}(k)\mathbf{H}(k))\mathbf{P}(k|k-1)(\mathbf{I} - \mathbf{K}(k)\mathbf{H}(k))^T + \mathbf{K}(k)\mathbf{R}(k)\mathbf{K}^T(k) \qquad (14)$$

## III. Kalman Filter Under MCC

*A. Derivation of the Algorithm*

Traditional Kalman filter works well under Gaussian noises, but its performance may deteriorate significantly under non-Gaussian noises, especially when the underlying system is disturbed by impulsive noises. The main reason for this is that KF is developed based on the MMSE criterion, which captures only the second order statistics of the error signal and is sensitive to large outliers. To address this problem, we propose in this work to use the MCC criterion to derive a new Kalman filter, which may perform much better in non-Gaussian noise environments, since correntropy contains second and higher order moments of the error.

For the linear model described in the previous section, we have

$$\begin{bmatrix} \hat{\mathbf{x}}(k|k-1) \\ \mathbf{y}(k) \end{bmatrix} = \begin{bmatrix} \mathbf{I} \\ \mathbf{H}(k) \end{bmatrix} \mathbf{x}(k) + \nu(k) \quad (15)$$

where $\mathbf{I}$ is the $n \times n$ identity matrix, and $\nu(k)$ is

$$\nu(k) = \begin{bmatrix} -(\mathbf{x}(k) - \hat{\mathbf{x}}(k|k-1)) \\ \mathbf{r}(k) \end{bmatrix},$$

with

$$\begin{aligned} &E\left[\nu(k)\nu^T(k)\right] \\ &= \begin{bmatrix} \mathbf{P}(k|k-1) & 0 \\ 0 & \mathbf{R}(k) \end{bmatrix} \\ &= \begin{bmatrix} \mathbf{B}_p(k|k-1)\mathbf{B}_p^T(k|k-1) & 0 \\ 0 & \mathbf{B}_r(k)\mathbf{B}_r^T(k) \end{bmatrix} \quad (16) \\ &= \mathbf{B}(k)\mathbf{B}^T(k) \end{aligned}$$

where $\mathbf{B}(k)$ can be obtained by Cholesky decomposition of $E\left[\nu(k)\nu^T(k)\right]$. Left multiplying both sides of (15) by $\mathbf{B}^{-1}(k)$, we get

$$\mathbf{D}(k) = \mathbf{W}(k)\mathbf{x}(k) + \mathbf{e}(k) \quad (17)$$

where $\mathbf{D}(k) = \mathbf{B}^{-1}(k) \begin{bmatrix} \hat{\mathbf{x}}(k|k-1) \\ \mathbf{y}(k) \end{bmatrix}$, $\mathbf{W}(k) = \mathbf{B}^{-1}(k) \begin{bmatrix} \mathbf{I} \\ \mathbf{H}(k) \end{bmatrix}$, $\mathbf{e}(k) = \mathbf{B}^{-1}(k)\nu(k)$. Since $E\left[\mathbf{e}(k)\mathbf{e}^T(k)\right] = \mathbf{I}$, the residual error $\mathbf{e}(k)$ are white.

Now we propose the following MCC based cost function:

$$J_L(x) = \frac{1}{L}\sum_{i=1}^{L} G_\sigma\left(d_i(k) - \mathbf{w}_i(k)\mathbf{x}(k)\right) \quad (18)$$

where $d_i(k)$ is the $i$-th element of $\mathbf{D}(k)$, $\mathbf{w}_i(k)$ is the $i$-th row of $\mathbf{W}(k)$, and $L = n + m$ is the dimension of $\mathbf{D}(k)$. Then, under MCC criterion, the optimal estimate of $\mathbf{x}(k)$ is

$$\hat{\mathbf{x}}(k) = \arg\max_{x} J_L(x) = \arg\max_{x} \sum_{i=1}^{L} G_\sigma\left(e_i(k)\right) \quad (19)$$

where $e_i(k)$ is the $i$-th element of $\mathbf{e}(k)$:

$$e_i(k) = d_i(k) - \mathbf{w}_i(k)\mathbf{x}(k) \quad (20)$$

The optimal solution can thus be obtained by solving

$$\frac{\partial J_L(x)}{\partial \mathbf{x}(k)} = 0 \quad (21)$$

It follows easily that

$$\mathbf{x}(k) = \left(\sum_{i=1}^{L}\left[G_\sigma\left(e_i(k)\right)\mathbf{w}_i^T(k)\mathbf{w}_i(k)\right]\right)^{-1} \times \\ \left(\sum_{i=1}^{L}\left[G_\sigma\left(e_i(k)\right)\mathbf{w}_i^T(k)d_i(k)\right]\right) \quad (22)$$

Since $e_i(k) = d_i(k) - \mathbf{w}_i(k)\mathbf{x}(k)$, the optimal solution (22) is actually a fixed-point equation of $\mathbf{x}(k)$ and can be rewritten as

$$\mathbf{x}(k) = f\left(\mathbf{x}(k)\right) \quad (23)$$

with

$$f\left(\mathbf{x}(k)\right) = \left(\sum_{i=1}^{L}\left[G_\sigma\left(d_i(k) - \mathbf{w}_i(k)\mathbf{x}(k)\right)\mathbf{w}_i^T(k)\mathbf{w}_i(k)\right]\right)^{-1} \times \\ \left(\sum_{i=1}^{L}\left[G_\sigma\left(d_i(k) - \mathbf{w}_i(k)\mathbf{x}(k)\right)\mathbf{w}_i^T(k)d_i(k)\right]\right)$$

A fixed-point iterative algorithm can be readily obtained as

$$\hat{\mathbf{x}}(k)_{t+1} = f\left(\hat{\mathbf{x}}(k)_t\right) \quad (24)$$

where $\hat{\mathbf{x}}(k)_t$ denotes the solution at the fixed-point iteration $t$.

The fixed-point equation (22) can also be expressed as

$$\mathbf{x}(k) = \left(\mathbf{W}^T(k)\mathbf{C}(k)\mathbf{W}(k)\right)^{-1}\mathbf{W}^T(k)\mathbf{C}(k)\mathbf{D}(k) \quad (25)$$

where $\mathbf{C}(k) = \begin{bmatrix} \mathbf{C}_x(k) & 0 \\ 0 & \mathbf{C}_y(k) \end{bmatrix}$, with

$\mathbf{C}_x(k) = diag\left(G_\sigma\left(e_1(k)\right), ..., G_\sigma\left(e_n(k)\right)\right)$,
$\mathbf{C}_y(k) = diag\left(G_\sigma\left(e_{n+1}(k)\right), ..., G_{n+m}\left(e_{n+m}(k)\right)\right)$.

The equation (25) can be further expressed as follows (see the Appendix for a detailed derivation):

$$\mathbf{x}(k) = \hat{\mathbf{x}}(k|k-1) + \overline{\mathbf{K}}(k)\left(\mathbf{y}(k) - \mathbf{H}(k)\hat{\mathbf{x}}(k|k-1)\right) \quad (26)$$

where

$$\begin{cases} \overline{\mathbf{K}}(k) = \overline{\mathbf{P}}(k|k-1)\mathbf{H}^T(k)\left(\mathbf{H}(k)\overline{\mathbf{P}}(k|k-1)\mathbf{H}^T(k)+\overline{\mathbf{R}}(k)\right)^{-1} \\ \overline{\mathbf{P}}(k|k-1) = \mathbf{B}_p(k|k-1)\mathbf{C}_x^{-1}(k)\mathbf{B}_p^T(k|k-1) \\ \overline{\mathbf{R}}(k) = \mathbf{B}_r(k)\mathbf{C}_y^{-1}(k)\mathbf{B}_r^T(k) \end{cases} \quad (27)$$

*Remark*: Of course, the equation (26) is also a fixed-point equation of $\mathbf{x}(k)$ because $\overline{\mathbf{K}}(k)$ depends on $\overline{\mathbf{P}}(k|k-1)$ and $\overline{\mathbf{R}}(k)$, both related to $\mathbf{x}(k)$ via $\mathbf{C}_x(k)$ and $\mathbf{C}_y(k)$, respectively. The optimal solution of the equation (26) depends also on the prior estimate $\hat{\mathbf{x}}(k|k-1)$, which can be calculated by (10) using the latest estimate $\hat{\mathbf{x}}(k-1|k-1)$.

With the above derivations, we summarize the proposed MCKF algorithm as follows:

1) Choose a proper kernel bandwidth $\sigma$ and a small positive number $\varepsilon$; Set an initial estimate $\hat{\mathbf{x}}(0|0)$ and an initial covariance matrix $\mathbf{P}(0|0)$; Let $k=1$;
2) Use equations (10) (11) to obtain $\hat{\mathbf{x}}(k|k-1)$ and $\mathbf{P}(k|k-1)$, and use Cholesky decomposition to obtain $\mathbf{B}_p(k|k-1)$;
3) Let $t=1$ and $\hat{\mathbf{x}}(k|k)_0 = \hat{\mathbf{x}}(k|k-1)$, where $\hat{\mathbf{x}}(k|k)_t$ denotes the estimated state at the fixed-point iteration $t$;
4) Use (28)-(34) to compute $\hat{\mathbf{x}}(k|k)_t$;

$$\hat{\mathbf{x}}(k|k)_t = \hat{\mathbf{x}}(k|k-1) + \widetilde{\mathbf{K}}(k)\left(\mathbf{y}(k) - \mathbf{H}(k)\hat{\mathbf{x}}(k|k-1)\right) \quad (28)$$

with

$$\widetilde{\mathbf{K}}(k) = \widetilde{\mathbf{P}}(k|k-1)\mathbf{H}^T(k)\left(\mathbf{H}(k)\widetilde{\mathbf{P}}(k|k-1)\mathbf{H}^T(k)+\widetilde{\mathbf{R}}(k)\right)^{-1}, \quad (29)$$

$$\widetilde{\mathbf{P}}(k|k-1) = \mathbf{B}_p(k|k-1)\widetilde{\mathbf{C}}_x^{-1}(k)\mathbf{B}_p^T(k|k-1), \quad (30)$$

$$\widetilde{\mathbf{R}}(k) = \mathbf{B}_r(k)\widetilde{\mathbf{C}}_y^{-1}(k)\mathbf{B}_r^T(k). \quad (31)$$

$$\widetilde{\mathbf{C}}_x(k) = diag\left(G_\sigma\left(\tilde{e}_1(k)\right),...,G_\sigma\left(\tilde{e}_n(k)\right)\right) \quad (32)$$

$$\widetilde{\mathbf{C}}_y(k) = diag\left(G_\sigma\left(\tilde{e}_{n+1}(k)\right),...,G_{n+m}\left(\tilde{e}_{n+m}(k)\right)\right) \quad (33)$$

$$\tilde{e}_i(k) = d_i(k) - \mathbf{w}_i(k)\hat{\mathbf{x}}(k|k)_{t-1} \quad (34)$$

5) Compare the estimation of the current step and the estimation of the last step. If (35) holds, set $\hat{\mathbf{x}}(k|k) = \hat{\mathbf{x}}(k|k)_t$ and continue to 6). Otherwise, $t+1 \rightarrow t$, and go back to 4).

$$\frac{\left\|\hat{\mathbf{x}}(k|k)_t - \hat{\mathbf{x}}(k|k)_{t-1}\right\|}{\left\|\hat{\mathbf{x}}(k|k)_{t-1}\right\|} \le \varepsilon \quad (35)$$

6) Update the posterior covariance matrix by (36), $k+1 \rightarrow k$ and go back to 2).

$$\mathbf{P}(k|k) = \left(\mathbf{I} - \widetilde{\mathbf{K}}(k)\mathbf{H}(k)\right)\mathbf{P}(k|k-1)\left(\mathbf{I} - \widetilde{\mathbf{K}}(k)\mathbf{H}(k)\right)^T \\ + \widetilde{\mathbf{K}}(k)\mathbf{R}(k)\widetilde{\mathbf{K}}^T(k) \quad (36)$$

*Remark*: As one can see, different from the traditional KF algorithm, the MCKF uses a fixed-point algorithm to update the posterior estimate of the state. The small positive number $\varepsilon$ provides a stop condition (or a threshold) for the fixed-point iteration. Since the initial value of the fixed-point iteration is set at the prior estimate $\hat{\mathbf{x}}(k|k-1)$, the convergence to the optimal solution will be very fast (usually in several steps).

The bandwidth $\sigma$ is a key parameter in MCKF. In general, a smaller bandwidth makes the algorithm more robust (with respect to outliers) but converge more slowly. On the other hand, when $\sigma$ becomes more and more larger, the MCKF will behave more and more like the ordinary KF algorithm. In particular, the following theorem holds.

*Theorem 1*: When the kernel bandwidth $\sigma \rightarrow \infty$, the MCKF will reduce to the KF algorithm.

*Proof*: see Appendix.

B. *Computational Complexity*

Next, we analyze the computational complexity in terms of the floating point operations for the proposed algorithm. The computational complexities of some basic equations are given in Table I.

TABLE I
COMPUTATIONAL COMPLEXITIES OF SOME EQUATIONS

| equation | multiplication and addition/subtraction | division, matrix inversion, Cholesky decomposition and exponentiation |
|---|---|---|
| (10) | $2n^2-n$ | 0 |
| (11) | $4n^3-n^2$ | 0 |
| (12) | $4n^2m+4nm^2-3nm$ | $O(m^3)$ |
| (13) | $4nm$ | 0 |
| (14) | $4n^3+6n^2m-2n^2+2nm^2-nm$ | 0 |
| (28) | $4nm$ | 0 |
| (29) | $4n^2m+4nm^2-3nm$ | $O(m^3)$ |
| (30) | $2n^3$ | $n + O(n^3)$ |
| (31) | $2m^3$ | $m + O(m^3)$ |
| (32) | $2n^2$ | $n$ |
| (33) | $2nm$ | $m$ |
| (34) | $2n$ | 0 |
| (36) | $4n^3+6n^2m-2n^2+2nm^2-nm$ | 0 |

The traditional Kalman filter algorithm involves the equations (10)~(14). Thus from Table I, one can conclude that the computational complexity of Kalman filter is

$$S_{KF} = 8n^3 + 10n^2m - n^2 + 6nm^2 - n + O(m^3) \quad (37)$$

The MCKF algorithm mainly involves the equations (10), (11), (28)~(34) and (36). Note that $\widetilde{\mathbf{C}}_x(k)$ and $\widetilde{\mathbf{C}}_y(k)$ are diagonal matrices, so it is very easy to obtain their inverse matrices. Assume that the average fixed-point iteration number is $T$. Then, according to Table I, the computational complexity of the MCKF is

$$S_{MCKF} = (2T+8)n^3 + (6+4T)Tn^2m + (2T-1)n^2 + (4T+2)nm^2 \quad (38)$$
$$+ (3T-1)nm + (4T-1)n + 2Tm^3 + 2Tm + TO(n^3) + 2TO(m^3)$$

The fixed-point iteration number $T$ is relatively small in general (see the simulation results in the next section). Thus the computational complexity of the MCKF is moderate compared with the traditional KF algorithm.

### C. Convergence Issue

The rigorous convergence analysis of the proposed MCKF algorithm is very complicated. In the following, we present only a sufficient condition that guarantees the convergence of the fixed-point iterations in MCKF. The result is similar to that of [26] and hence, will not be proved here.

Let $\|\cdot\|_p$ denote an $l_p$-norm of a vector or an induced norm of a matrix defined by $\|\mathbf{A}\|_p = \max_{\|\mathbf{X}\|_p \neq 0} \frac{\|\mathbf{A}\mathbf{X}\|_p}{\|\mathbf{X}\|_p}$ with $p \geq 1$, and $\lambda_{\min}[.]$ denote the minimum eigenvalue of a matrix. According to the results of [26], the following theorem holds.

***Theorem 2***: If $\beta > \zeta = \dfrac{\sqrt{n}\sum_{i=1}^{L}\|\mathbf{w}_i^T(k)\|_1 |d_i(k)|}{\lambda_{\min}\left[\sum_{i=1}^{L}\mathbf{w}_i^T(k)\mathbf{w}_i(k)\right]}$, and

$\sigma \geq \max\{\sigma^*, \sigma^\dagger\}$, in which $\sigma^*$ is the solution of the equation $\phi(\sigma) = \beta$, with

$$\phi(\sigma) = \frac{\sqrt{n}\sum_{i=1}^{L}\|\mathbf{w}_i^T(k)\|_1 |d_i(k)|}{\lambda_{\min}\left[\sum_{i=1}^{L}G_\sigma\left(\beta\|\mathbf{w}_i(k)\|_1 + |d_i(k)|\right)\mathbf{w}_i^T(k)\mathbf{w}_i(k)\right]}, \sigma \in (0,\infty) \quad (39)$$

and $\sigma^\dagger$ is the solution of equation $\psi(\sigma) = \alpha (0 < \alpha < 1)$, with

$$\psi(\sigma) = \frac{\sqrt{n}\sum_{i=1}^{L}\left[\left(\beta\|\mathbf{w}_i(k)\|_1 + |d_i(k)|\right)\|\mathbf{w}_i(k)\|_1\left(\beta\|\mathbf{w}_i^T(k)\mathbf{w}_i(k)\|_1 + \|\mathbf{w}_i^T(k)d_i(k)\|_1\right)\right]}{\sigma^2 \lambda_{\min}\left[\sum_{i=1}^{L}G_\sigma\left(\beta\|\mathbf{w}_i(k)\|_1 + |d_i(k)|\right)\mathbf{w}_i^T(k)\mathbf{w}_i(k)\right]}$$

$$\sigma \in (0,\infty) \quad (40)$$

then it holds that $\|\mathbf{f}(\mathbf{x}(k))\|_1 \leq \beta$, and $\|\nabla_{\mathbf{x}(k)}\mathbf{f}(\mathbf{x}(k))\|_1 \leq \alpha$ for all $\mathbf{x}(k) \in \{\mathbf{x}(k) \in \mathbb{R}^n : \|\mathbf{x}(k)\|_1 \leq \beta\}$, where $\nabla_{\mathbf{x}(k)}\mathbf{f}(\mathbf{x}(k))$ denotes the $n \times n$ Jacobian matrix of $\mathbf{f}(\mathbf{x}(k))$ with respect to $\mathbf{x}(k)$, that is

$$\nabla_{\mathbf{x}(k)}\mathbf{f}(\mathbf{x}(k)) = \left[\frac{\partial}{\partial x_1(k)}\mathbf{f}(\mathbf{x}(k))\ldots\frac{\partial}{\partial x_n(k)}\mathbf{f}(\mathbf{x}(k))\right] \quad (41)$$

with

$$\frac{\partial}{\partial x_j(k)}\mathbf{f}(\mathbf{x}(k))$$
$$= -\mathbf{N}_{ww}^{-1}\left(\frac{1}{\sigma^2}\sum_{i=1}^{L}\left[e_i(k)w_i^j(k)G_\sigma(e_i(k))\mathbf{w}_i^T(k)\mathbf{w}_i(k)\right]\right)\mathbf{f}(\mathbf{x}(k))$$
$$+ \mathbf{N}_{ww}^{-1}\left(\frac{1}{\sigma^2}\sum_{i=1}^{L}\left[e_i(k)w_i^j(k)G_\sigma(e_i(k))\mathbf{w}_i^T(k)d_i(k)\right]\right)$$

where $\mathbf{N}_{ww} = \sum_{i=1}^{L}G_\sigma(e_i(k))\mathbf{w}_i^T(k)\mathbf{w}_i(k)$ and $w_i^j(k)$ is the $j$-th element of $\mathbf{w}_i(k)$.

By Theorem 2, if the kernel bandwidth $\sigma$ is larger than a certain value, we have

$$\begin{cases} \|\mathbf{f}(\mathbf{x}(k))\|_1 \leq \beta \\ \|\nabla_{\mathbf{x}(k)}\mathbf{f}(\mathbf{x}(k))\|_1 \leq \alpha < 1 \end{cases} \quad (42)$$

By *Banach Fixed-Point Theorem* [24], given an initial state estimate satisfying $\|\mathbf{x}(k)_0\|_1 \leq \beta$, the fixed-point iteration algorithm in MCKF will surely converge to a unique fixed point in the range $\mathbf{x}(k) \in \{\mathbf{x}(k) \in \mathbb{R}^n : \|\mathbf{x}(k)\|_1 \leq \beta\}$ provided that the kernel bandwidth $\sigma$ is larger than a certain value (e.g. $\max\{\sigma^*, \sigma^\dagger\}$).

Theorem 2 implies that the kernel bandwidth has significant influence on the convergence behavior of MCKF. If the kernel bandwidth $\sigma$ is too small, the algorithm will diverge or converge very slowly. A larger kernel bandwidth ensures a faster converge speed but usually leads to a poor performance in impulsive noises. In practical applications, the bandwidth can be set manually or optimized by *trial and error* methods.

## IV. ILLUSTRATIVE EXAMPLES

In this section, we present two illustrative examples to demonstrate the performance of the proposed MCKF algorithm, and compare it to the traditional KF algorithm.

### A. Example 1

Consider the following linear system:

$$\begin{bmatrix} x_1(k) \\ x_2(k) \end{bmatrix} = \begin{bmatrix} \cos(\theta) & -\sin(\theta) \\ \sin(\theta) & \cos(\theta) \end{bmatrix}\begin{bmatrix} x_1(k-1) \\ x_2(k-1) \end{bmatrix} + \begin{bmatrix} q_1(k-1) \\ q_2(k-1) \end{bmatrix} \quad (43)$$

$$y(k) = \begin{bmatrix} 1 & 1 \end{bmatrix}\begin{bmatrix} x_1(k) \\ x_2(k) \end{bmatrix} + r(k) \quad (44)$$

where $\theta = \pi/18$.

First, we consider the case in which the noises are all Gaussian, that is

$$q_1(k-1) \sim N(0, 0.01)$$
$$q_2(k-1) \sim N(0, 0.01)$$
$$r(k) \sim N(0, 0.01)$$

Table II shows the MSEs of $x_1$ and $x_2$ for different filters. Here the MSE is computed as an average over 100 independent Monte Carlo runs, and in each run, 1000 samples (time steps) are used to evaluate the MSE. Since all the noises are Gaussian, the Kalman filter performs very well and in this example, it achieves almost the best performance (that is, the smallest MSEs). One can also see that when the kernel bandwidth is too small, the MCKF may achieve a worse performance; while when the bandwidth becomes larger, its performance will approach that of the KF. Actually, it has been proved that when $\sigma \to \infty$, the MCKF will reduce to the traditional KF. In general, one should choose a larger kernel width under Gaussian noises.

TABLE II
MSEs OF $x_1$ AND $x_2$ IN GAUSSIAN NOISES

| Filter | MSE of $x_1$ | MSE of $x_2$ |
| --- | --- | --- |
| KF | 0.035778 | 0.030052 |
| MCKF$(\sigma=0.5, \varepsilon=10^{-6})$ | 0.131361 | 0.105729 |
| MCKF$(\sigma=1.0, \varepsilon=10^{-6})$ | 0.103497 | 0.096126 |
| MCKF$(\sigma=3.0, \varepsilon=10^{-6})$ | 0.035885 | 0.030139 |
| MCKF$(\sigma=5.0, \varepsilon=10^{-6})$ | 0.035785 | 0.030047 |
| MCKF$(\sigma=7.0, \varepsilon=10^{-6})$ | 0.035784 | 0.030051 |

Second, we consider the case in which the process noises are still Gaussian but the observation noise is a heavy-tailed (impulsive) non-Gaussian noise, with a mixed-Gaussian distribution, that is

$$q_1(k-1) \sim N(0, 0.01)$$
$$q_2(k-1) \sim N(0, 0.01)$$
$$r(k) \sim 0.9N(0, 0.01) + 0.1N(0, 100)$$

Fig.1 and Fig.2 illustrate the probability densities of the estimation errors of $x_1$ and $x_2$. In the simulation, we set $\varepsilon = 10^{-6}$. As one can see, in impulsive noises, when kernel bandwidth is too small or too large, the performance of MCKF will be not good. In this case, however, with a proper kernel bandwidth (say $\sigma = 2.0$), the MCKF can outperform the KF significantly, achieving a desirable error distribution with a higher peak and smaller dispersion. Again, when $\sigma$ is very large, MCKF achieves almost the same performance as the KF.

Fig.3 shows the fixed-point iteration numbers at the time step (or instant) $k=1$ for different kernel bandwidths. It is evident that the larger the kernel bandwidth, the faster the convergence speed. In particular, when the kernel bandwidth is large enough, the fixed-point algorithm in MCKF will converge to the optimal solution in just one or two iterations. In practical applications, to avoid slow convergence, the kernel bandwidth cannot be set at a very small value. Similar results can also be seen from Table III, where the average fixed-point iteration numbers of every time step for different filters are shown, which are computed as averages over 100 independent Monte Carlo runs, with each run containing 1000 time steps.

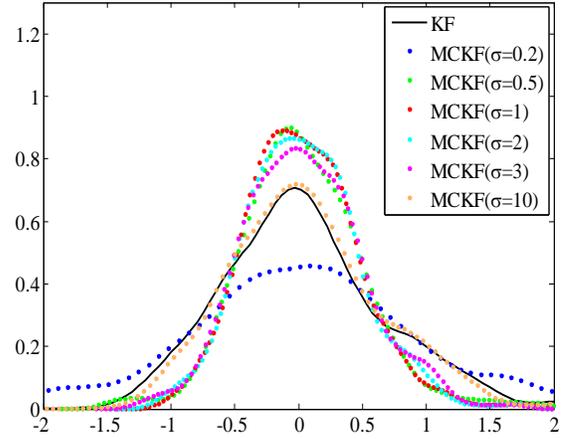

Fig. 1. Probability densities of $x_1$ estimation errors with different filters

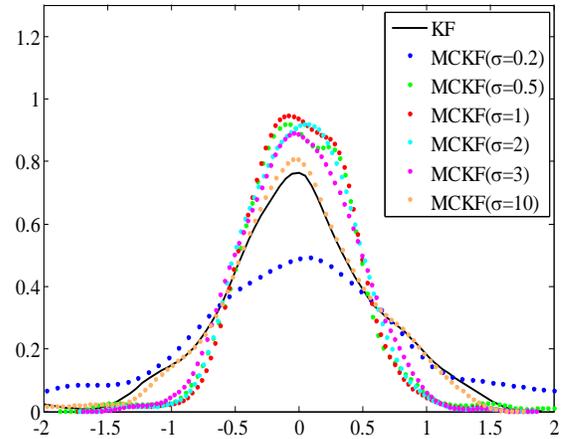

Fig. 2. Probability densities of $x_2$ estimation errors with different filters

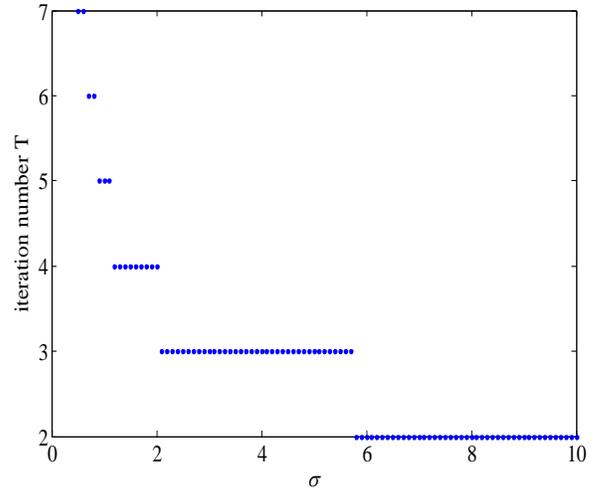

Fig. 3. Fixed-point iteration numbers at time step $k=1$ for different kernel bandwidths

TABLE III
AVERAGE ITERATION NUMBERS FOR EVERY TIME STEP WITH DIFFERENT $\sigma$

| Filter | Average Iteration Numbers |
|---|---|
| MCKF$(\sigma=0.2, \varepsilon=10^{-6})$ | 3.89826 |
| MCKF$(\sigma=0.5, \varepsilon=10^{-6})$ | 2.78835 |
| MCKF$(\sigma=1.0, \varepsilon=10^{-6})$ | 2.36406 |
| MCKF$(\sigma=2.0, \varepsilon=10^{-6})$ | 2.12967 |
| MCKF$(\sigma=3.0, \varepsilon=10^{-6})$ | 2.01343 |
| MCKF$(\sigma=10, \varepsilon=10^{-6})$ | 1.66423 |

We further investigate the influence of the threshold $\varepsilon$ on the performance. Table IV illustrates the MSEs of $x_1$ and $x_2$ with different $\varepsilon$ (where the kernel bandwidth is set at $\sigma=2.0$), and Table V presents the average fixed-point iteration numbers. One can see that a smaller $\varepsilon$ usually results in slightly lower MSEs but needs more iterations to converge. Obviously, the influence of $\varepsilon$ is not significant compared with the kernel bandwidth $\sigma$.

TABLE IV
MSEs OF $x_1$ AND $x_2$ WITH DIFFERENT $\varepsilon$

| Filter | MSE of $x_1$ | MSE of $x_2$ |
|---|---|---|
| MCKF$(\sigma=2.0, \varepsilon=10^{-1})$ | 0.221182 | 0.168721 |
| MCKF$(\sigma=2.0, \varepsilon=10^{-2})$ | 0.220386 | 0.167958 |
| MCKF$(\sigma=2.0, \varepsilon=10^{-4})$ | 0.220326 | 0.167900 |
| MCKF$(\sigma=2.0, \varepsilon=10^{-6})$ | 0.220322 | 0.167899 |
| MCKF$(\sigma=2.0, \varepsilon=10^{-8})$ | 0.220322 | 0.167899 |

TABLE V
AVERAGE ITERATION NUMBERS FOR EVERY TIME STEP WITH DIFFERENT $\varepsilon$

| Filter | Average Iteration Numbers |
|---|---|
| MCKF$(\sigma=2.0, \varepsilon=10^{-1})$ | 1.03884 |
| MCKF$(\sigma=2.0, \varepsilon=10^{-2})$ | 1.09134 |
| MCKF$(\sigma=2.0, \varepsilon=10^{-4})$ | 1.44313 |
| MCKF$(\sigma=2.0, \varepsilon=10^{-6})$ | 2.13328 |
| MCKF$(\sigma=2.0, \varepsilon=10^{-8})$ | 2.78392 |

Fig.4 and Fig.5 show the true and the estimated values of $x_1(k)$ and $x_2(k)$ with KF and MCKF ($\sigma=2.0, \varepsilon=10^{-6}$). The results clearly indicate that the MCKF can achieve much better tracking performance than the traditional KF algorithm.

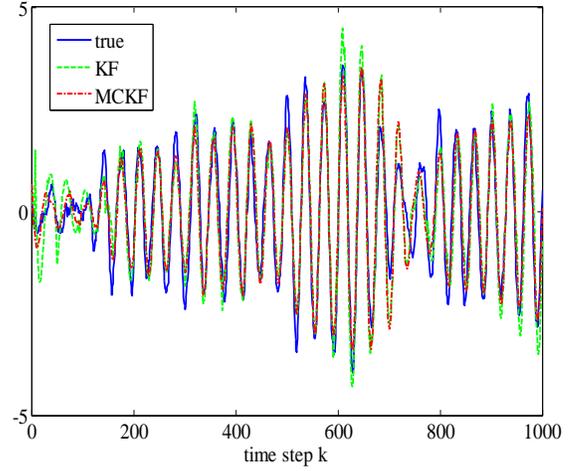

Fig. 4. The true and the estimated values of $x_1$

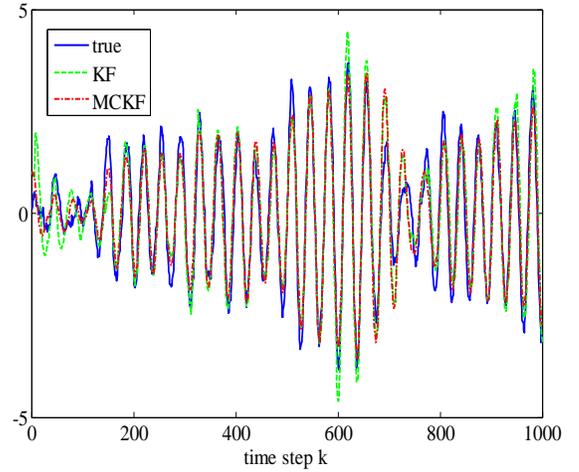

Fig. 5. The true and the estimated values of $x_2$

*B. Example 2*

Now we consider a practical example about one-dimensional linear uniformly accelerated motion. The state vector is $\mathbf{x}(k)=[x_1(k)\ x_2(k)\ x_3(k)]^T$, in which $x_1(k)$ is the position, $x_2(k)$ denotes the speed, and $x_3(k)$ stands for the acceleration. We assume that there are certain noises in the system and only the speed can be observed, which is also affected by some measurement disturbances. $\Delta T$ represents the measurement time interval. Then, the state and measurement equations are given by

$$\begin{bmatrix} x_1(k) \\ x_2(k) \\ x_3(k) \end{bmatrix} = \begin{bmatrix} 1 & \Delta T & 0 \\ 0 & 1 & \Delta T \\ 0 & 0 & 1 \end{bmatrix} \begin{bmatrix} x_1(k-1) \\ x_2(k-1) \\ x_3(k-1) \end{bmatrix} + \begin{bmatrix} q_1(k-1) \\ q_2(k-1) \\ q_3(k-1) \end{bmatrix} \quad (45)$$

$$y(k) = \begin{bmatrix} 0 & 1 & 0 \end{bmatrix} \begin{bmatrix} x_1(k) \\ x_2(k) \\ x_3(k) \end{bmatrix} + r(k) \quad (46)$$

with $\Delta T = 0.1s$. First, the process noises are assumed to be Gaussian and the measurement noise is non-Gaussian with a mixed-Gaussian distribution, that is

$$q_1(k-1) \sim N(0, 0.01)$$
$$q_2(k-1) \sim N(0, 0.01)$$
$$q_3(k-1) \sim N(0, 0.01)$$
$$r(k) \sim 0.9N(0, 0.01) + 0.1N(0, 100)$$

Further, the initial values of the true state, estimated state and covariance matrix are assumed to be:

$$\mathbf{x}(0) = [0 \ \ 0 \ \ 1]^T,$$
$$\hat{\mathbf{x}}(0|0) = [0 \ \ 0 \ \ 1]^T + N(0, 0.01) \times [1 \ \ 1 \ \ 1]^T,$$
$$\mathbf{P}(0|0) = 0.01 \times diag\{1 \ \ 1 \ \ 1\}.$$

Fig.6~ Fig. 8 demonstrate the probability densities of the estimation errors of $x_1$, $x_2$ and $x_3$ for KF and MCKF, and Table VI summarizes the corresponding MSEs. In the simulation, the parameters are set at $\sigma = 2.0, \varepsilon = 10^{-6}$. Those results confirm again that the proposed MCKF can outperform the traditional KF significantly when the system is disturbed by Gaussian process noises and non-Gaussian measurement noise.

TABLE VI
MSEs OF $x_1$, $x_2$ AND $x_3$ IN GAUSSIAN PROCESS NOISES AND NON-GAUSSIAN MEASUREMENT NOISE

| Filter | MSE of $x_1$ | MSE of $x_2$ | MSE of $x_3$ |
|---|---|---|---|
| KF | $50.7874 \, (m^2)$ | $0.8172 \, (m^2/s^2)$ | $0.2719 \, (m^2/s^4)$ |
| MCKF | $10.1444 \, (m^2)$ | $0.3133 \, (m^2/s^2)$ | $0.2231 \, (m^2/s^4)$ |

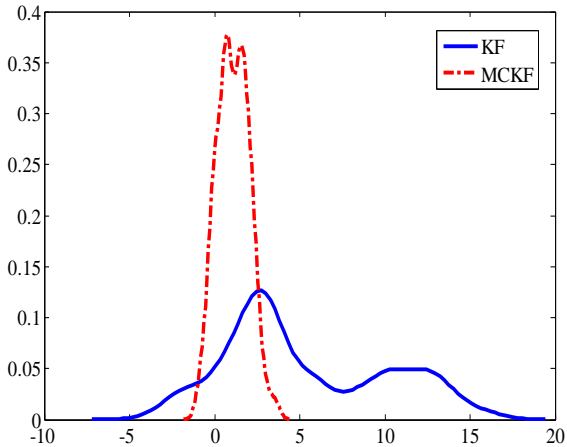

Fig. 6. Probability densities of $x_1$ estimation errors for KF and MCKF in Gaussian process noises and non-Gaussian measurement noise

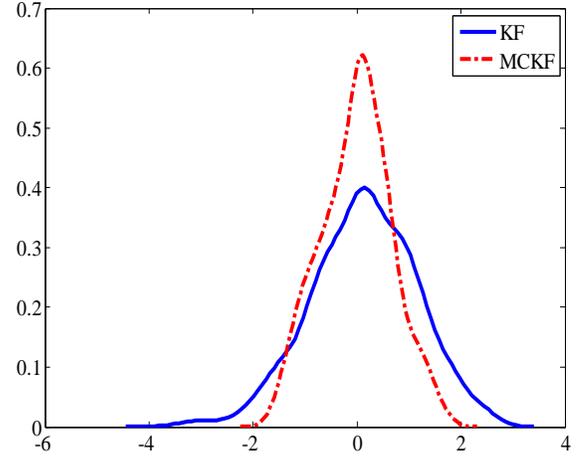

Fig. 7. Probability densities of $x_2$ estimation errors for KF and MCKF in Gaussian process noises and non-Gaussian measurement noise

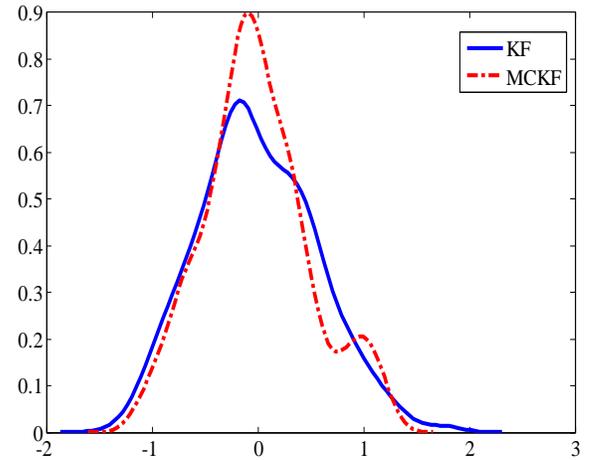

Fig. 8. Probability densities of $x_3$ estimation errors for KF and MCKF in Gaussian process noises and non-Gaussian measurement noise

Next, we consider the situation where the process and measurement noises are all non-Gaussian with mixed-Gaussian distributions, that is

$$q_1(k-1) \sim 0.9N(0, 0.01) + 0.1N(0, 1)$$
$$q_2(k-1) \sim 0.9N(0, 0.01) + 0.1N(0, 1)$$
$$q_3(k-1) \sim 0.9N(0, 0.01) + 0.1N(0, 1)$$
$$r(k) \sim 0.9N(0, 0.01) + 0.1N(0, 100)$$

With the same initial values and parameters setting as before, the results are shown in Fig.9~11 and Table VII. As expected, the MCKF performs much better than the traditional KF when the system is disturbed by non-Gaussian process and measurement noises.

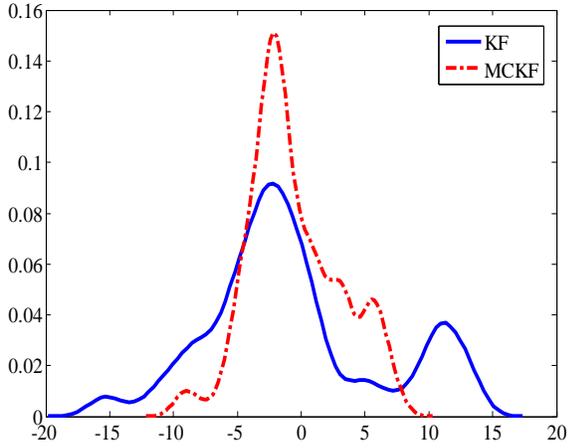

Fig. 9. Probability densities of $x_1$ estimation errors for KF and MCKF in non-Gaussian process and measurement noises

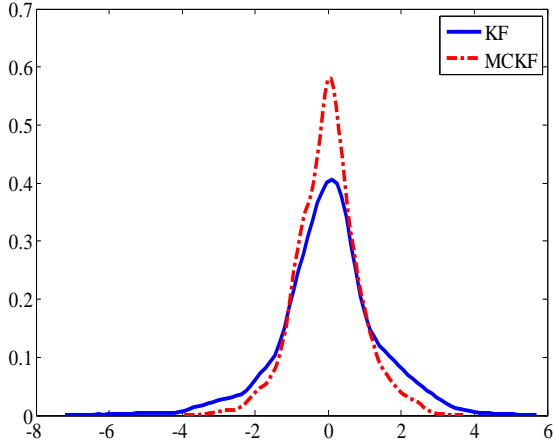

Fig. 10. Probability densities of $x_2$ estimation errors for KF and MCKF in non-Gaussian process and measurement noises

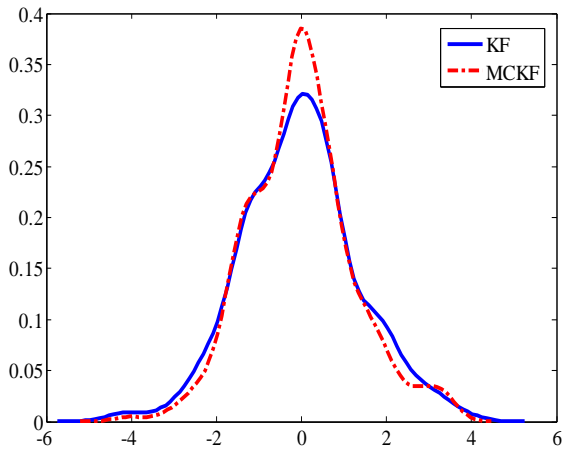

Fig. 11. Probability densities of $x_3$ estimation errors for KF and MCKF in non-Gaussian process and measurement noises

TABLE VII
MSEs OF $x_1$, $x_2$ AND $x_3$ IN NON-GAUSSIAN PROCESS AND MEASUREMENT NOISES

| Filter | MSE of $x_1$ | MSE of $x_2$ | MSE of $x_3$ |
|---|---|---|---|
| KF | $114.8233\,(m^2)$ | $1.6358\,(m^2/s^2)$ | $1.8149\,(m^2/s^4)$ |
| MCKF | $44.1290\,(m^2)$ | $0.7229\,(m^2/s^2)$ | $1.5803\,(m^2/s^4)$ |

## V. CONCLUSION

A new Kalman type filtering algorithm, called *maximum correntropy Kalman filter* (MCKF), has been proposed in this work. The MCKF is derived by using the *maximum correntropy criterion* (MCC) as the optimality criterion, instead of using the well-known *minimum mean square error* (MMSE) criterion. The propagation equations for the prior estimation of the state and covariance matrix in MCKF are the same as those in KF. However, different from the KF, the MCKF uses a novel fixed-point algorithm to update the posterior estimations. The computational complexity of the MCKF is not expensive and the convergence is ensured if the kernel bandwidth is larger than a certain value. When the kernel bandwidth is large enough, the MCKF will behave like the KF. With a proper kernel bandwidth, the MCKF can outperform the KF significantly, especially when the underlying system is disturbed by some impulsive non-Gaussian noises.

## APPENDIX

*A. Derivation of the formula (26)*

$$\mathbf{W}(k) = \mathbf{B}^{-1}(k) \begin{bmatrix} \mathbf{I} \\ \mathbf{H}(k) \end{bmatrix}$$

$$= \begin{bmatrix} \mathbf{B}_p^{-1}(k|k-1) & 0 \\ 0 & \mathbf{B}_r^{-1}(k) \end{bmatrix} \begin{bmatrix} \mathbf{I} \\ \mathbf{H}(k) \end{bmatrix}$$

$$= \begin{bmatrix} \mathbf{B}_p^{-1}(k|k-1) \\ \mathbf{B}_r^{-1}(k)\mathbf{H}(k) \end{bmatrix} \quad (A.1)$$

$$\mathbf{C}(k) = \begin{bmatrix} \mathbf{C}_x(k) & 0 \\ 0 & \mathbf{C}_y(k) \end{bmatrix} \quad (A.2)$$

$$\mathbf{D}(k) = \mathbf{B}^{-1}(k) \begin{bmatrix} \hat{\mathbf{x}}(k|k-1) \\ \mathbf{y}(k) \end{bmatrix}$$

$$= \begin{bmatrix} \mathbf{B}_p^{-1}(k|k-1)\hat{\mathbf{x}}(k|k-1) \\ \mathbf{B}_r^{-1}(k)\mathbf{y}(k) \end{bmatrix} \quad (A.3)$$

By (A.1) and (A.2), we have

$$\begin{aligned}&\left(\mathbf{W}^T(k)\mathbf{C}(k)\mathbf{W}(k)\right)^{-1}\\&=\left[\left(\mathbf{B}_p^{-1}\right)^T\mathbf{C}_x\mathbf{B}_p^{-1}+\mathbf{H}^T\left(\mathbf{B}_r^{-1}\right)^T\mathbf{C}_y\mathbf{B}_r^{-1}\mathbf{H}\right]^{-1}\end{aligned} \quad (A.4)$$

where we denote $\mathbf{B}_p(k|k-1)$ by $\mathbf{B}_p$, $\mathbf{B}_r(k)$ by $\mathbf{B}_r$, $\mathbf{C}_x(k)$ by $\mathbf{C}_x$ and $\mathbf{C}_y(k)$ by $\mathbf{C}_y$ for simplicity.

Using the *matrix inversion lemma* with the identification:

$$\left(\mathbf{B}_p^{-1}\right)^T\mathbf{C}_x\mathbf{B}_p^{-1}\to\mathbf{A},\ \mathbf{H}^T\to\mathbf{B},$$
$$\mathbf{H}\to\mathbf{C},\ \left(\mathbf{B}_r^{-1}\right)^T\mathbf{C}_y\mathbf{B}_r^{-1}\to\mathbf{D}$$

We arrive at

$$\begin{aligned}&\left(\mathbf{W}^T(k)\mathbf{C}(k)\mathbf{W}(k)\right)^{-1}\\&=\left(\mathbf{B}_p\mathbf{C}_x^{-1}\mathbf{B}_p^T-\mathbf{B}_p\mathbf{C}_x^{-1}\mathbf{B}_p^T\mathbf{H}^T(\mathbf{B}_r\mathbf{C}_y^{-1}\mathbf{B}_r^T+\mathbf{H}\mathbf{B}_p\mathbf{C}_x^{-1}\mathbf{B}_p^T\mathbf{H}^T)^{-1}\mathbf{H}\mathbf{B}_p\mathbf{C}_x^{-1}\mathbf{B}_p^T\right)\end{aligned} \quad (A.5)$$

Further, by (A.1)~(A.3), we derive

$$\begin{aligned}&\mathbf{W}^T(k)\mathbf{C}(k)\mathbf{D}(k)\\&=\left(\mathbf{B}_p^{-1}\right)^T\mathbf{C}_x\mathbf{B}_p^{-1}\hat{\mathbf{x}}(k|k-1)+\mathbf{H}^T\left(\mathbf{B}_r^{-1}\right)^T\mathbf{C}_y\mathbf{B}_r^{-1}\mathbf{y}(k)\end{aligned} \quad (A.6)$$

Combining (25), (A.5) and (A.6), we obtain (26).

*B. Proof of Theorem 1*

$$\lim_{\sigma\to\infty}\mathrm{G}_\sigma^{-1}\left(\tilde{e}_i(k)\right)=\lim_{\sigma\to\infty}\left[\exp\left(-\frac{\tilde{e}_i^2(k)}{2\sigma^2}\right)\right]^{-1}=1 \quad (A.7)$$

It follows easily that

$$\begin{aligned}\lim_{\sigma\to\infty}\widetilde{\mathbf{C}}_x^{-1}(k)&=\lim_{\sigma\to\infty}diag\left(\mathrm{G}_\sigma^{-1}\left(\tilde{e}_1(k)\right),...,\mathrm{G}_\sigma^{-1}\left(\tilde{e}_n(k)\right)\right)\\&=diag(\underbrace{1,...,1}_{n})\end{aligned} \quad (A.8)$$

$$\begin{aligned}\lim_{\sigma\to\infty}\widetilde{\mathbf{C}}_y^{-1}(k)&=\lim_{\sigma\to\infty}diag\left(\mathrm{G}_\sigma^{-1}\left(\tilde{e}_{n+1}(k)\right),...,\mathrm{G}_\sigma^{-1}\left(\tilde{e}_{n+m}(k)\right)\right)\\&=diag(\underbrace{1,...,1}_{m})\end{aligned} \quad (A.9)$$

$$\begin{aligned}\lim_{\sigma\to\infty}\widetilde{\mathbf{P}}(k|k-1)&=\lim_{\sigma\to\infty}\mathbf{B}_p(k|k-1)\widetilde{\mathbf{C}}_x^{-1}(k)\mathbf{B}_p^T(k|k-1)\\&\stackrel{(A.8)}{=}\mathbf{B}_p(k|k-1)\mathbf{B}_p^T(k|k-1)\\&\stackrel{(16)}{=}\mathbf{P}(k|k-1)\end{aligned} \quad (A.10)$$

$$\begin{aligned}\lim_{\sigma\to\infty}\widetilde{\mathbf{R}}(k)&=\lim_{\sigma\to\infty}\mathbf{B}_r(k)\widetilde{\mathbf{C}}_y^{-1}(k)\mathbf{B}_r^T(k)\\&\stackrel{(A.9)}{=}\mathbf{B}_r(k)\mathbf{B}_r^T(k)\\&\stackrel{(16)}{=}\mathbf{R}(k)\end{aligned} \quad (A.11)$$

$$\begin{aligned}\lim_{\sigma\to\infty}\widetilde{\mathbf{K}}(k)&=\lim_{\sigma\to\infty}\left[\widetilde{\mathbf{P}}(k|k-1)\mathbf{H}^T(k)\left(\mathbf{H}(k)\widetilde{\mathbf{P}}(k|k-1)\mathbf{H}^T(k)+\widetilde{\mathbf{R}}(k)\right)^{-1}\right]\\&\stackrel{(A.10)(A.11)}{=}\mathbf{P}(k|k-1)\mathbf{H}^T(k)\left(\mathbf{H}(k)\mathbf{P}(k|k-1)\mathbf{H}^T(k)+\mathbf{R}(k)\right)^{-1}\\&\stackrel{(12)}{=}\mathbf{K}(k)\end{aligned} \quad (A.12)$$

$$\begin{aligned}\lim_{\sigma\to\infty}\hat{\mathbf{x}}(k|k)_t&=\lim_{\sigma\to\infty}\left[\hat{\mathbf{x}}(k|k-1)+\widetilde{\mathbf{K}}(k)\left(\mathbf{y}(k)-\mathbf{H}(k)\hat{\mathbf{x}}(k|k-1)\right)\right]\\&\stackrel{(A.12)}{=}\hat{\mathbf{x}}(k|k-1)+\mathbf{K}(k)\left(\mathbf{y}(k)-\mathbf{H}(k)\hat{\mathbf{x}}(k|k-1)\right)\\&\stackrel{(13)}{=}\hat{\mathbf{x}}(k|k)\end{aligned} \quad (A.13)$$

$$\begin{aligned}&\lim_{\sigma\to\infty}\widetilde{\mathbf{P}}(k|k)\\&=\lim_{\sigma\to\infty}\left[\left(\mathbf{I}-\widetilde{\mathbf{K}}(k)\mathbf{H}(k)\right)\mathbf{P}(k|k-1)\left(\mathbf{I}-\widetilde{\mathbf{K}}(k)\mathbf{H}(k)\right)^T+\widetilde{\mathbf{K}}(k)\mathbf{R}(k)\widetilde{\mathbf{K}}^T(k)\right]\\&\stackrel{(A.12)}{=}\left(\mathbf{I}-\mathbf{K}(k)\mathbf{H}(k)\right)\mathbf{P}(k|k-1)\left(\mathbf{I}-\mathbf{K}(k)\mathbf{H}(k)\right)^T+\mathbf{K}(k)\mathbf{R}(k)\mathbf{K}^T(k)\\&\stackrel{(14)}{=}\mathbf{P}(k|k)\end{aligned} \quad (A.14)$$

This completes the proof.